\documentclass[lettersize,journal]{IEEEtran}
\usepackage{amsmath,amsfonts}
\usepackage{algorithmic}
\usepackage{algorithm}
\usepackage{array}
\usepackage[caption=false,font=normalsize,labelfont=sf,textfont=sf]{subfig}
\usepackage{textcomp}
\usepackage{stfloats}
\usepackage{url}
\usepackage{verbatim}
\usepackage{graphicx}
\usepackage{cite}
\usepackage{upgreek}
\usepackage{multirow}

\begin{document}

\title{Geometric Multimodal Deep Learning with Multi-Scaled Graph Wavelet Convolutional Network}

\author{Maysam~Behmanesh,~\IEEEmembership{Graduate~Student~Member,~IEEE,}
        Peyman~Adibi,
        Mohammad~Saeed~Ehsani,
        and~Jocelyn~Chanussot,~\IEEEmembership{Fellow,~IEEE}

\thanks{Maysam Behmanesh, Peyman Adibi (corresponding author) and Mohammad Saeed Ehsani are with Artificial Intelligence Department, Faculty of Computer Engineering, University of Isfahan, Iran e-mail: (mbehmanesh@eng.ui.ac.ir, adibi@eng.ui.ac.ir, ehsani@eng.ui.ac.ir)}
\thanks{Jocelyn Chanussot is with University of Grenoble Alpes, CNRS, Grenoble INP, GIPSA-lab, 38000 Grenoble, France e-mail(jocelyn.chanussot@gipsa-lab.grenoble-inp.fr)}
\thanks{Manuscript received December --, 2021; revised --, 2022.}}

%

\maketitle

\begin{abstract}
Multimodal data provide complementary information of a natural phenomenon by integrating data from various domains with very different statistical properties. Capturing the intra-modality and cross-modality information of multimodal data is the essential capability of multimodal learning methods. The geometry-aware data analysis approaches provide these capabilities by implicitly representing data in various modalities based on their geometric underlying structures. Also, in many applications, data are explicitly defined on an intrinsic geometric structure. Generalizing deep learning methods to the non-Euclidean domains is an emerging research field, which has recently been investigated in many studies. Most of those popular methods are developed for unimodal data. In this paper, a multimodal multi-scaled graph wavelet convolutional network (M-GWCN) is proposed as an end-to-end network. M-GWCN simultaneously finds intra-modality representation by applying the multiscale graph wavelet transform to provide helpful localization properties in the graph domain of each modality, and cross-modality representation by learning permutations that encode correlations among various modalities. M-GWCN is not limited to either the homogeneous modalities with the same number of data, or any prior knowledge indicating correspondences between modalities. Several semi-supervised node classification experiments have been conducted on three popular unimodal explicit graph-based datasets and five multimodal implicit ones. The experimental results indicate the superiority and effectiveness of the proposed methods compared with both spectral graph domain convolutional neural networks and state-of-the-art multimodal methods.

\end{abstract}


\section{Introduction}
\label{sec1.introduction}

Multimodal data are generally composed of several heterogeneous data gathered from multiple real-world phenomena. Different modalities provide parts of the description of a phenomenon from various source domains usually with very different statistical properties. Unlike the unimodal data that may indicate only partial information of an entity, multimodal data provide complementary information by leveraging and fusing modality-specific knowledge \cite{Baltrusaitis2019}. Although each modality has its distinct statistical properties, the modalities are usually semantically correlated. Multimodal learning tries to improve the performance in applications with multimodal data by discovering the hidden intra-modality and cross-modality correlations.\par

To apply the geometry-aware data analysis for multimodal problems, data in various modalities can be implicitly represented based on their geometric structures such as graphs, manifolds, and meshed surfaces. Also, there are an increasing number of applications, in which data are generated in non-Euclidean geometries and inherently defined for example as a graph. These applications represent complex relationships and interdependencies among objects \cite{Wu2021}, including social networks, citation networks, networks of the spread of epidemic diseases, e-commerce networks, brain’s neuronal networks, biological regulatory networks, and so on.\par

Due to the increasing growth of non-Euclidean data with explicit or implicit geometric structures, and success of deep learning models in capturing hidden patterns in various domains, many deep learning methods have recently been revolutionized to be applied on geometrically structured domains. \par

Graph neural networks (GNNs) are the most common networks developing deep learning methods for graphs as the geometric structure of data, which perform filtering operations directly on the graph via the graph weights \cite{7974879}.
Graph convolutional network (GCN) can learn the local meaningful stationary properties of the input signals through especially designed convolution operator on graphs \cite{Scarselli2009}. GCN represents node features (also called node embeddings) within these local neighborhoods on the graph to learn graph embedding for node classification, graph signal classification, graph regression, and so on \cite{Scarselli2009}.
Complex geometric structures in graphs can be encoded with the strong mathematical tools in many spatial or spectral graph-based methods \cite{Chung2001}. In spatial graph-based methods, the convolution on each vertex is directly defined by aggregating feature information from all its neighbors \cite{Micheli2009}. Spectral graph-based methods define convolution by leveraging graph Fourier transform to convert signals defined in vertex domain into the spectral domain using convolution theorem \cite{JoanBruna2014}. \par

Most of the recent multimodal data analysis methods are limited because are based on two simplifying assumptions: 1) there is the same number of data samples in each modality (the homogeneity assumption), and 2) at least partial correspondences and/or non-correspondences between modalities are given as prior knowledge. Obviously, these assumptions are unrealistic in practical machine learning scenarios \cite{Behmanesh2021CrossModalAM}.\par

The success of GCNs for unimodal graph-based data and the limitations of current multimodal methods in practical problems motivates the approach proposed in this paper, that is extending spectral GCN models to multimodal graph-based data. Although various spectral and spatial methods are developed to extend convolutional neural networks on graph-based data, there are not numerous methods for extending GCNs for multimodal graph-based domains.\par

The main purpose of this work is to introduce a novel spectral GCN on graph-based multimodal data in a more practical scenario, in which it is only known that the data points of various modalities are sampled from the same phenomena. \par

In this paper, we proposed a multimodal graph wavelet convolutional network (M-GWCN), with an end-to-end learning. Unlike existing spectral graph-based methods that apply graph Fourier transforms for providing localization properties of the spectral domain, M-GWCN uses graph wavelet transforms as a multiresolution analysis of graph signals. Besides the capability of M-GWCN in highly localizing in the vertex domain, it can simultaneously localize signal content in both spatial and spectral domains (Graph wavelets from the spectral graph theory point of view are studied in \cite{Hammond2011}). \par

To prevent over-smoothing, M-GWCN is equipped with an initial residual connection mechanism. It makes sure that the representation at each layer contains a portion of the initial features. To avoid the expensive spectral decomposition for diagonalizing the graph Laplacian, we apply the Chebyshev polynomial method for approximating the graph wavelet bases. This approximation enables the M-GWCN to be applicable for large graphs.\par

The proposed M-GWCN consists of two main parts. In the first part, a graph convolution is conducted by applying multi-scaled graph wavelet transforms in each modality. Graph wavelet bases obtained with different scales provide valuable localization properties in graph domain of each modality. These properties are applied for intra-modality representation of feature vectors of each graph.  In the second part, cross-modality representations are computed in one layer, that is responsible for data fusion. In this layer, permutations for encoding the cross-modality correlations are learned by applying new loss function and regularization terms.\par

The main contributions of this paper are briefly summarized as follows:\par

\begin{enumerate}
\item We generalize the spectral GCN model to multimodal graph-based data domains as a problem that has rarely been addressed.
\item We consider a general scenario for multimodal problems with unpaired data in the heterogeneous modalities, which needs no prior knowledge.
\item We introduce a new spectral GCN model that in parallel applies graph wavelet transforms with different scaling parameters in each modality. This method prevents over-smoothing by introducing initial residual connections.
\item We develop a new efficient network with end-to-end learning that simultaneously applies feature mapping on each modality and explores permutations for representing the cross-modality correlations among various modalities.
\item We experimentally demonstrate the superiority and effectiveness of the proposed M-GWCN model on three unimodal explicit graph-based datasets and five multimodal datasets, which are not explicitly defined as graphs, but the graphs are constructed based on them implicitly.
\end{enumerate}

The remainder of the paper is organized as follows. Related works are reviewed in section \ref{sec2.relatedWorks}. In section \ref{sec3.bakground}, we overview the background and the basic notations. Our proposed methods are introduced in section \ref{sec4.proposedMethod}. Experimental results are presented and discussed in section \ref{sec5.experiments}. Finally, section \ref{sec6.conclusions} concludes the paper.\par

\section{Related works}
\label{sec2.relatedWorks}

\subsection{Multimodal learning}
\label{sec2A.multimodalLearning}

Many multimodal learning methods are limited to the problems with homogeneous data, which have the same number of data samples in all modalities. For example, MVML-LA is a multi-view method, which learns a common discriminative low-dimensional latent space by preserving the geometric structure of the original input \cite{Zhao2018}. MVDL-CV is a new multi-view method, which obtains the sparse representation of the sample by learning a particular dictionary for each view and determines the similarity of samples using a regularization term between two dictionaries \cite{LIU2021157}. GPLVM represents multiple modalities in a common subspace using the Gaussian process latent variable model \cite{LI2021108}. Another work provides a unifying framework for multiclass classification by encompassing vector-valued manifold regularization and co-regularized multi-view learning \cite{Minh2016}. \par

Some multimodal learning methods are geometry-aware and try to extend different diffusion and spectral methods to the multimodal setting. A nonlinear multimodal data fusion method captures the intrinsic structure of data and relies on minimal prior model knowledge \cite{Katz2019}. A multimodal image registration method uses the graph Laplacian to capture the intrinsic structure of data in each modality by introducing a new structure preservation criterion based on Laplacian commutativity \cite{Zimmer2019}. \par

These two methods assume that the data samples in various modalities are completely paired, and all modalities have the same number of data samples. Another approach presented in \cite{Eynard2015}, assumed partial correspondence information is predetermined using an expert manipulation process.\par

Since the correspondence knowledge between various modalities, which is essential for the above learning multimodal methods, may not be available in many practical scenarios, working independent of this prior knowledge is of key importance. Thus, it has been focused in continue on several methods, which try to reduce dependency on the expert manipulation process.\par

The method presented in \cite{Pournemat2021} first extends the given correspondence information between modalities using functional mapping idea on the data manifolds of the respected modalities and then uses all correspondence information to simultaneously learn an underlying low-dimensional common manifold by aligning the manifolds of different modalities. In \cite{Behmanesh2021}, another multimodal manifold learning approach, called local signal expansion for joint diagonalization (LSEJD) was proposed, which uses the intrinsic local tangent spaces to expand the initial correspondences knowledge. \par

Although these two later methods greatly expand correspondence information, they still depend on the little basic prior knowledge of correspondences. In recent work \cite{Behmanesh2021CrossModalAM}, we proposed a multimodal learning method, which is independent from any prior expertise between modalities. This method first uses spectral graph wavelet transform for representing local descriptors of each modality, and then applies these descriptors to find point-wise correspondences between modalities using the functional map approach.\par

\subsection{Graph convolutional neural networks}
\label{sec2B.graphConvolutionalNeuralNetworks}

Inspired by the success of CNNs in the Euclidean domain, a large number of methods are proposed to generalize CNNs to non-Euclidean and especially graph domains. These methods are classified into two categories, spatial methods and spectral ones. \par

Spatial methods directly perform convolution on a graph based on its topological structure. Different spatial methods provide various weighted average functions for characterizing the node influences in the neighborhoods. NN4G \cite{Micheli2009} performs graph convolutions by summing up the nodes neighborhood information directly. The message-passing neural network (MPNN) \cite{Gilmer2017} treats graph convolution as a message-passing process in which information can be passed from one node to another along edges directly. To amend the MPNN-based methods in distinguishing different graph structures based on the graph embedding, the graph isomorphism network (GIN) \cite{Xu2019a} is proposed. GIN adjusts the weight of the central node in a neighborhood by a learnable parameter. GraphSage \cite{Hamilton2017} performs graph convolution by adopting a sampling strategy to obtain a fixed number of neighbors for each node. Graph attention network (GAT) \cite{Velickovic2018} applies graph convolution by adopting attention mechanisms to learn the relative weights between two connected nodes. \par

Spectral methods developed the graph convolution operation based on the spectral graph theory, which is generally based on Fourier analysis in the graph domain. Spectral CNN (SCNN) \cite{JoanBruna2014} is the first spectral method, which its graph convolution layer projects the input graph signal to a new space using graph Fourier transform. To amend the limitations of the basic SCNN, Chebyshev spectral CNN (ChebyNet) \cite{Defferrard2016} is introduced, which applies a fast localized filter approximation to find desired approximate filter response through the Chebyshev expansion. CayleyNet is another spectral method that uses Cayley polynomials filters. Unlike ChebyNet, CayleyNet can detect narrow frequency bands with a small number of filter parameters \cite{Levie2019}. To address the limitations of spectral graph CNN methods in providing beneficial localization properties, GWNN is presented that applies graph wavelets as a set of bases instead of eigenvectors of graph Laplacian \cite{Xu2019}. A new spectral method is proposed in \cite{Bianchi2021} that offers a more flexible frequency response. It captures a better global graph structure by applying the Auto-Regressive Moving Average (ARMA) filter.\par

\subsection{Multimodal graph neural networks}
\label{sec2C.multimodalGraphNeuralNetworks}

Generalizing GNN to graph-based multimodal data is an important problem that has rarely been addressed. A multimodal GNN method for visual question answering tasks is proposed \cite{Gao2020}. This method represents the image as three graph-based modalities and refines the features of nodes by passing a message from one graph to another.  Inspired by the message-passing idea of graph neural networks, a multimodal GCN (MMGCN) framework is proposed in \cite{Wei2019}. MMGCN captures user preferences in a recommender system by enriching the representation of each node by leveraging information interchange in various modalities. Motivated by graph-based structure in addressing the long unaligned sequences, a multimodal GCN-based method is proposed in \cite{Mai2020} to investigate the effectiveness of GNN in modeling the multimodal sequential data. The Edge Adaptable GCN (EA-GCN) method for disease prediction is presented in \cite{Huang2020}. EA-GCN represents various modalities as a population graph using an edge adapter and applies GCN for semi-supervised node classification. \par

\section{Background}
\label{sec3.bakground}

\subsection{Multimodal problem formulation}
\label{sec3.1.multimodalProblemFormulation}

We assume that multimodal data are acquired from $M$ different modalities or spaces $\left\{ \mathcal{M}_m\right\}_{m=1}^M$ with different dimensions $\left\{d_m\right\}_{m=1}^M$. Data in each modality $m$ is represented by an undirected weighted graph $\mathcal{G}_m=(\mathcal{V}_m,\mathcal{E}_m,\mathbf{A}_m)$, where $\mathcal{V}_m$ is a set of $N_m$ vertices, $\mathcal{E}_m$ is the set of $E_m$ edges, and $\mathbf{A}_m\in \mathbb{R}^{N_m\times N_m}$ is a weighted adjacency matrix representing connection weights between vertices. \par

Let graph signal $\mathbf{X}_m=\left[\mathbf{x}_{m,1}...\mathbf{x}_{m,{N_m}} \right]^T \in \mathbb{R}^{N_m\times d_m}$ is a collection of all feature vectors associated with $l_m$ labeled and $u_m$ unlabeled vertices, where $N_m=l_m+u_m$.\par

Consider $\mathbf{Y}_m =\left[\mathbf{y}_{m,1} ... \mathbf{y}_{m,{N_m}}\right]^T\in \left\{0,1\right\}^{N_m\times C}$ as the label matrix of data sample in modality $m$ for a $C$-class classification problem. If the vertex $\mathbf{x}_{m,j}$ belongs to the $k$-th class, then $\mathbf{y}_{m,j}$ contains $1$ in the $k$-th location and $0$ in all others. For unlabeled data sample $\mathbf{x}_{m,k}$, the vector $\mathbf{y}_{m,k}$ has $-1$ in all $C$ locations. \par

In this paper, it is assumed that the sample correspondences information across different modalities is unknown. Also, when there is no emphasis on a specific modality, symbols are considered in the generic notation, e.g., $\mathcal{G}$ is used to show the data graph instead of $\mathcal{G}_m$, that indicates a specific modality $m$. \par

\subsection{Graph spectral geometry}
\label{sec3.2.graphSpectralGeometry}

The symmetric normalized Laplacian matrix of graph $\mathcal{G}_m$ is defined as  $\mathbf{L}_m=\mathbf{D}_m^{-1/2} (\mathbf{D}_m-\mathbf{A}_m)\mathbf{D}_m^{-1/2}$, by discretizing the Laplace-Beltrami (LB) operator \cite{Rosenberg1997}, where $\mathbf{D}_m=diag(\sum_{k\neq l} w_{k,l}^m)$ is the diagonal matrix of nodes degrees. For all $M$ Laplacian matrices $\left\{\mathbf{L}_m \in \mathbb{R}^{N_m\times N_m} \right\}_{m=1}^M$, there are unitary eigenspaces $\mathbf{U}^m$’s, such that $\mathbf{L}_m=\mathbf{U}_m \mathbf{\Lambda}_m \mathbf{U}_m^T$, where matrix $\mathbf{U}_m=\left[\mathbf{u}_{m,1}... \mathbf{u}_{m,{N_m}}\right]$ and diagonal matrix $\mathbf{\Lambda}_m = diag(\mathbf{\lambda}_{m,1}, ... , \mathbf{\lambda}_{m,{N_m}})$ contains the orthogonal eigenvectors and their corresponding eigenvalues (spectrum), respectively. \par

The eigenvectors $\mathbf{U}_m$ play the role of Fourier basis in classical harmonic analysis, and the eigenvalues $\mathbf{\Lambda}_m$ can be interpreted as frequencies. For a given graph signal $\mathbf{f}=(f_{m,1},...,f_{m,{N_m}})^T \in \mathbb{R}^{N_m}$ on the vertices of $\mathcal{G}_m$, $\mathbf{\hat{f}}=\mathbf{U}_m^T \mathbf{f}$ performs the graph Fourier transform, and $\mathbf{f}=\mathbf{U}_m \mathbf{\hat{f}}$ is its inverse.

\subsection{Spectral graph convolutional network}
\label{sec3.3.spectralGraphConvolutionalNetwork}

According to convolution theorem, the spectral graph convolution of signal $\mathbf{x}$ with the filter $\mathbf{g}\in \mathbb{R}^N$ on the graph $\mathcal{G}$ can be defined as an element-wise product of their Fourier transform as $\mathbf{x}\star \mathbf{g}=\mathbf{U} ((\mathbf{U}^T \mathbf{x})\odot(\mathbf{U}^T \mathbf{g}))$. By denoting $\mathbf{g}_{\theta}=diag(\mathbf{U}^T \mathbf{g})$, the spectral graph convolution can be written in the form of matrix multiplication as $\mathbf{x}*\mathbf{g}_{\theta}=\mathbf{U} \mathbf{g}_{\theta} \mathbf{U}^T \mathbf{x}$ \cite{Defferrard2016}.\par

The spectral convolution layer $k$ is defined by extending CNN to graph $\mathcal{G}$ as follows \cite{JoanBruna2014}:

\begin{equation}
\label{eq1}
\begin{aligned}
\mathbf{X}^{k}(:,j)=&\sigma\left(\sum_{i=1}^{f_{k-1}} \left( \sum_{r=1}^R \uptheta_{i,j}^{(k)}(\mathbf{\lambda}_r) \mathbf{u}_r \mathbf{u}_r^T \right)\mathbf{X}^{k-1}(:,i) \right)=\\
&\sigma\left(\sum_{i=1}^{f_{k-1}} \mathbf{U}_R \mathbf{\uptheta}_{i,j}^{(k)}(\mathbf{\Lambda}_R)\mathbf{U}_R^T  \mathbf{X}^{k-1}(:,i) \right),\\
&i=1, ...,f_k,
\end{aligned}
\end{equation}

\noindent where $\mathbf{U}_R$ is an $N\times R$ matrix of first $R$ eigenvectors of Laplacian matrix $L$ (corresponding to its least eigenvalues), $\mathbf{g}_{\uptheta}=\mathbf{\theta}_{i,j}^{(k)} (\mathbf{\Lambda}_R)$ is an $R\times R$ diagonal matrix of spectral multipliers representing the learnable parameters in layer $k$, $\mathbf{X}^{(k-1)}=\left[\mathbf{X}^{(k-1)}(:,1)... \mathbf{X}^{(k-1)}(:,f_{k-1})\right] \in \mathbb{R}^{N\times f_{k-1}}$ is the input signal including $f_{k-1}$ features (channels), $\mathbf{X}^{(k)}=\left[\mathbf{X}^{(k)} (:,1)... \mathbf{X}^\mathbf{(k)}(:,f_k)\right]\in \mathbb{R}^{N\times f_k}$ is the output signal including $f_k$ features, and $\sigma(\cdot)$ is a nonlinear activation function (e.g., ReLU). In this equation, parameter $R$ keeps the locality of filter in the spectral domain using the $R$ lowest frequency harmonies. \par

The main drawback of spectral CNN is its computational complexity because the eigen-decomposition problem of Laplacian matrix $\mathbf{L}$ is too expensive, and also yields dense eigenvectors $\mathbf{U}$ that disables taking advantage of sparse multiplications. \par

To overcome this limitation, ChebyNet \cite{Defferrard2016} model is presented that approximates the desired filter response $\mathbf{g}_\theta$ using Chebyshev expansion as follows:

\begin{equation}
\label{eq2}
\mathbf{g}_\theta\boldsymbol=\sum^R_{i=0}\theta _i  \mathbf{T}_i(\widetilde {\mathbf{\Lambda}}),
\end{equation}

\noindent where $\mathbf{\widetilde{\Lambda}}=2\mathbf{\Lambda}/\mathbf{\lambda}_{max} - \mathbf{I}_N$ is the rescaled spectrum in $[-1,1]$, $\uptheta=\left[\theta_0 ... \theta_R \right]$ is the $(R+1)$-dimensional vector of the polynomial coefficients parametrizing the filter and is optimized during the training, and $\mathbf{T}_i (\mathbf{\widetilde{\Lambda}})=2 \mathbf{\widetilde{\Lambda}} \mathbf{T}_{i-1} (\mathbf{\widetilde{\Lambda}})-\mathbf{T}_{i-2} (\mathbf{\widetilde{\Lambda}})$ is the Chebyshev polynomial of order $i$ that defines recursively with  $\mathbf{T}_0 (\mathbf{\widetilde{\Lambda}})=1$ and $\mathbf{T}_1 (\mathbf{\widetilde{\Lambda}})=\mathbf{\widetilde{\Lambda}} $.\par

The convolution of graph signal $\mathbf{x}$ with this defined filter $\mathbf{g}_\theta$ is obtained as follows: \par

\begin{equation}
\label{eq3}
\mathbf{x}\mathbf{*}{\mathbf{g}}_{\theta }\mathbf{=}\mathbf{U}\left(\sum^R_{i=0}{{\theta }_i\ }{\mathbf{T}}_i\left(\widetilde{\mathbf{\Lambda }}\right)\right){\mathbf{U}}^T\mathbf{x}=\sum^R_{i=0}{{\theta }_i\ }{\mathbf{T}}_i\left(\widetilde{\mathbf{L}}\right)\mathbf{x},
\end{equation}

\noindent where $\mathbf{\widetilde{L}}=2\mathbf{L}/\mathbf{\lambda}_{max}-\mathbf{I}_N$. The resulting convolution layer $k$ is now define as: \par

\begin{equation}
\label{eq4}
\mathbf{X}^{(k)}=\sigma \left(\sum^R_{i=0}\mathbf{T}_i\left(\widetilde{\mathbf{L}}\right)\mathbf{X}^{(k-1)}\mathbf{W}^{(k)}_i\right),
\end{equation}

\noindent where $\mathbf{W}_i^{(k)}\in \mathbb{R}^{f_{k-1}\times {f_k}}$ indicates $i$-th trainable weight matrix in layer $k$.\par

A specific polynomial order $i$ in equation (\ref{eq4}) covers the $i$-hope neighborhood and ignores the impact of the farther neighbors. To cover the larger structures in graph, it is essential to apply a high-order polynomial, but this polynomial leads to overfitting to the known graph. Furthermore, this high-order polynomial is more computationally expensive. \par

Graph Convolutional Network (GCN) \cite{Kipf2017} presents a simplified version of the Chebyshev filter. It reduces complexity and overfitting by setting  $R=1$, $\mathbf{\lambda}_{max}=2$, $\mathbf{W}_0=-\mathbf{W}_1=\mathbf{W}$, and substituting $\mathbf{\widetilde{L}}$ by $\mathbf{\widehat{A}}=\mathbf{\widetilde{D}}^{-1/2} \mathbf{\widetilde{A}} \mathbf{\widetilde{D}} ^{-1/2}$ in equation (\ref{eq4}), where $\mathbf{\widetilde{A}}=\mathbf{A}+\mathbf{I}$ and $\mathbf{\widetilde{D}}_{ii}=\sum_j \mathbf{\widetilde{A}}_{ij}$. Thus, the convolution layer $k$ of GCN is obtained as:

\begin{equation}
\label{eq5}
\mathbf{X}^{(k)}=\sigma\left(\widehat{\mathbf{A}}{\mathbf{X}}^{(k-1)}{\mathbf{W}}^{(k)}\right).
\end{equation}

This model is able to cover the large structure of graph with high-order neighborhoods by applying multiple GCN layers. \par

The other limitations, such as representing narrow-band filter with ChebyNet and smoothing node features after few convolutions with GCN, have been addressed using Cayley \cite{Levie2019} and ARMA filters \cite{Bianchi2021}. The rational forms of these filters can offer a large variety of shapes for them.

\subsection{Spectral graph wavelet transform}
\label{sec3.4.spectralGraphWaveletTransform}

Wavelet transform is a powerful multiresolution analysis tool that expresses a signal as a combination of several localized, shifted, and scaled bases (wavelet bases) \cite{Mallat2009}. Spectral graph wavelet transform refers to projecting graph signals from vertex domain into the spectral domain using a proper set of bases provided by wavelet transform. This transformation provides valuable localization property by applying a series of appropriate scaling operations of graph signals \cite{Hammond2011}. \par

As initially shown in \cite{Hammond2011}, the spectral graph wavelet localized at vertex $i$ with scale parameter $s$ is shown by $\uppsi_{s,i}$ whose $j$-th element is given by:

\begin{equation}
\label{eq6}
\mathbf{\uppsi}_{s,i}\left(j\right)=\sum^N_{l=1}{g\left(s{\lambda }_l\right)u^*_l(i)u_l(j)}, 
\end{equation}

\noindent where $N$ is the number of vertices, $\mathbf{\lambda}_l$ is the $l$-th eigenvalue of the normalized graph Laplacian matrix, $\mathbf{u}_l$ is the Laplacian’s associated eigenvector that its $j$-th element $u_l(j)$ is the value of the Laplace-Beltrami operator eigenfunction at vertex $j$, the symbol $*$ denotes complex conjugate operator, and $g:\mathbb{R}^{+}\rightarrow \mathbb{R}^{+}$ is the spectral graph wavelet generating kernel. \par

Wavelet bases are defined by $\mathbf{\Psi}_s=\left[\mathbf{\uppsi}_{s,1}... \mathbf{\uppsi}_{s,N} \right]\in \mathbb{R}^{N\times N}$, where each wavelet basis $\mathbf{\uppsi}_{s,i}$ corresponds to a signal on vertex  $i$ and scale $s$. According to wavelet bases, equation (\ref{eq6}) can be written in the form of matrix multiplication:

\begin{equation}
\label{eq7}
\mathbf{\Psi}_s=\mathbf{U} \mathbf{G}_s \mathbf{U}^T,
\end{equation}

\noindent where $\mathbf{G}_s=diag(g(s\mathbf{\lambda}_1),...,g(s\mathbf{\lambda}_N))$ is the scaling matrix and $g(s\mathbf{\lambda}_i)=e^{\mathbf{\lambda}_i s}$. Graph wavelet transform for a given graph signal $\mathbf{f}=(f_{i,1},...,f_{i,{N_i}})^T\in \mathbb{R}^{N_i}$ is define by $\mathbf{\hat{f}}=\mathbf{\Psi}_s^{-1} \mathbf{f}$, and its inverse is $\mathbf{f}=\mathbf{\Psi}_s \mathbf{\hat{f}}$. \par

According to \cite{Donnat2018}, $\mathbf{\Psi}_s^{-1}$ can be obtained by replacing $g(s\mathbf{\lambda}_i)$ in $\mathbf{\Psi}_s$ with $g(-s\mathbf{\lambda}_i)$ corresponding to a heat kernel. \par

Computing the wavelet bases is still dependent on eigen-decomposition, which is inefficient for large graphs. As mentioned in \cite{Hammond2011}, the graph wavelet bases can be approximated using Chebyshev polynomials as follows:

\begin{equation}
\label{eq8}
\begin{aligned}
\mathbf{\Psi }_s=&\frac{1}{2}c_{0,s}+\sum^Q_{i=1}{c_{i,s}\mathbf{T}}_i(\widetilde{\mathbf{L}}),\\
&c_{i,s}=2e^{-s}J_i(-s),
\end{aligned}
\end{equation}

\noindent where $\mathbf{T}_i(\widetilde{\mathbf{L}})=2\widetilde{\mathbf{L}}\mathbf{T}_{i-1}(\widetilde{\mathbf{L}})-\mathbf{T}_{i-2}(\widetilde{\mathbf{L}}) $ is the Chebyshev polynomial of order $i$ for approximate $\mathbf{\Psi}_s$ with $\mathbf{T}_{0}(\widetilde{\mathbf{L}})=1$, $\mathbf{T}_{1}(\widetilde{\mathbf{L}})=\widetilde{\mathbf{L}}$, $Q$ is the number of Chebyshev polynomials, and $J_i (-s)$ is the Bessel function of the first kind \cite{Arfken2013}.

\section{Proposed method}
\label{sec4.proposedMethod}

As mentioned in section \ref{sec3.bakground}, computational complexity of graph Fourier transform for obtaining Fourier bases is the main drawback of the most spectral methods. In addition, graph Fourier transform, as a global transformation, does not provide helpful localization properties in the vertex domain. \par

Inspired by the superiority of spectral graph wavelet transforms in approximating with highly sparse wavelet bases which have more helpful localization properties, we developed a novel multimodal graph wavelet convolution network (M-GWCN) for analyzing multimodal data.\par

Due to the sparsity of wavelet bases, the computations in the M-GWCN model that are based on wavelet bases are much more efficient than models, which are based on graph Fourier bases. Furthermore, since each wavelet basis is related to a signal on the graph that diffused away from a central node, the M-GWCN model with small scale values is localized in the vertex domain of each modality. Thus, different scales of wavelet bases enable this model to represent feature vectors of each modality based on the different levels of localities in an efficient way. \par

The M-GWCN model also takes advantage of the complementary information provided by various modalities by learning cross-modal representations. Cross-modality representation is the process of representing feature vectors of each modality based on wavelet bases of the other modalities. Finding the correspondences among different vertices in various graphs is essential for cross-modality representation. The M-GWCN model explores these correspondences by learning permutation matrices that encode them among various modalities.\par

The proposed M-GWCN method has three advantages distinguished it from previous networks: \par

\begin{enumerate}
\item M-GWCN introduces a stacked architecture that utilizes graph wavelets convolution with multiple scaling parameters in parallel and provides more useful intra-modality localization properties by aggregating embedded features at different scales.
    
\item M-GWCN adopts residual connections that not only are helpful to prevent the over-smoothing of each stack, but also encourage them to provide various filtering responses based on each scale.
    
\item M-GWCN generalizes the benefit of graph wavelet transform for cross-modality representation by finding permutations encoded cross-modality correlations among various modalities.

\end{enumerate}

\subsection{Multiscale adaptive graph wavelet}
\label{4.1.multiscaleAdaptiveGraphWavelet }

We define an Adaptive Graph Wavelet (AGW) as a building block of the proposed network. In each layer, AGW consists of a graph wavelet transformation with a desired scale and a residual connection as an adaptive component. AGWs with different scales are concatenated in parallel to form a multiscale adaptive graph wavelet (MAGW) or a stack of AGWs. Multiple scales of wavelets provide more helpful localization properties by decomposing a graph signal on components at different scales, or frequency ranges. \par

\begin{figure*}[!t]
\centering
\includegraphics[width=5in]{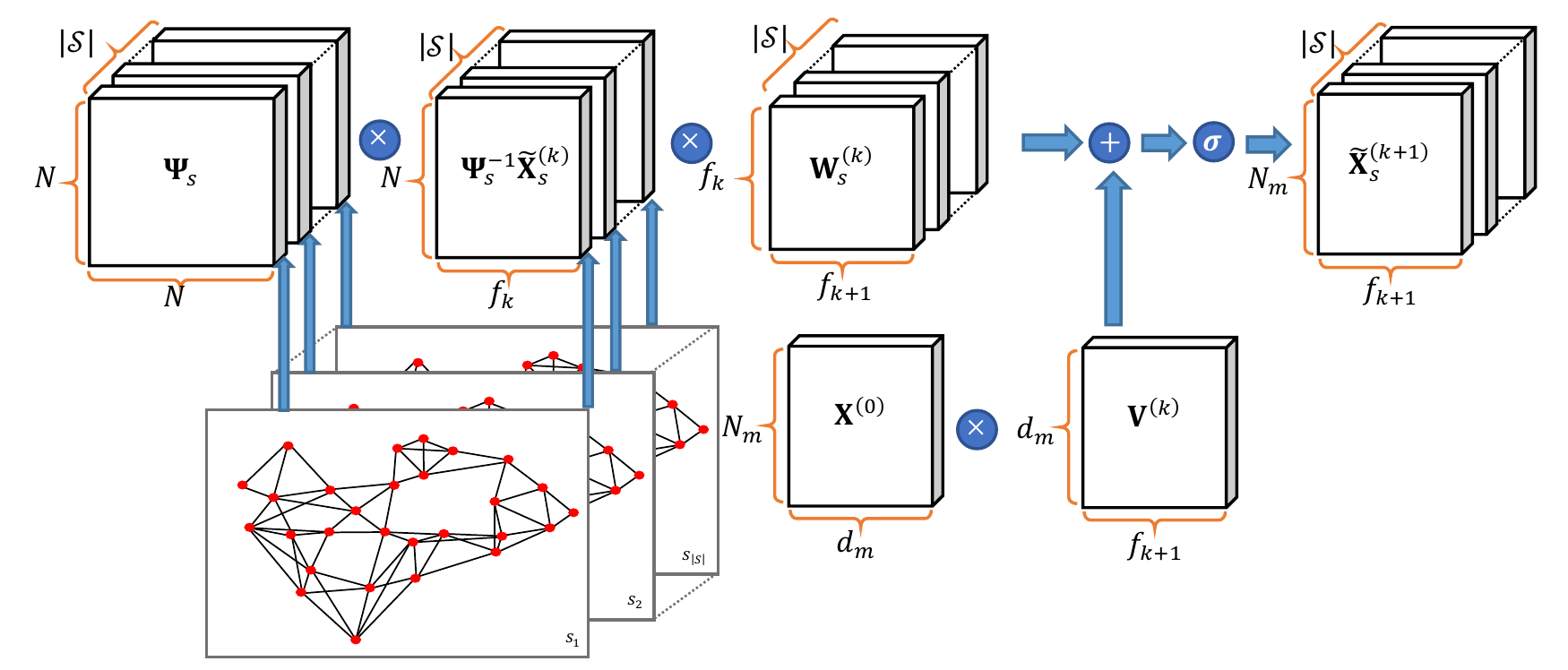}
\caption{Scheme of a stack of AGWs in layer $k$ with $\mid\mathcal{S}\mid$ scales. Each AGW consists of graph wavelet bases with a desired scale for feature mapping and a residual connection, as an adaptive component, for preventing the over-smoothing of each stack. AGWs with different scales are concatenated to form a multiscale AGW (MAGW) or a stack of AGWs.}
\label{fig1}
\end{figure*}

The designed graph filter with MAGW approximates graph frequency response with different scales, without knowing the underlying graph structure. The residual connection compensates over-smoothing of MAGW in deeper layers and provides different filtering responses for each AGW. Fig. \ref{fig1} depicts a scheme of MAGW. \par

According to the superiority of rational filters in approximating various shapes of filters, compared with polynomial filters, we apply a rational filter, as a more versatile graph filter, for each AGW.
Inspired by frequency response of rational filters mentioned in \cite{Isufi2017}, the response of filtering signal $\mathbf{x}$ in scale s can be implemented using the following first-order recursion:

\begin{equation}
\label{eq9}
\mathbf{x}^{(k)}_s=\omega_s \mathbf{M}\mathbf{x}^{(k-1)}_s+{\varphi }_s\mathbf{x}^{(0)}_s,
\end{equation}

\noindent where $\omega_s$ and $\varphi_s$ are the filter coefficients in scale $s$, and $\mathbf{M}$ is any practical graph representing matrix used to capture comprehensive information of the graph.\par

Inspired by this filter response in graph signal processing, we design a machine learning approach for learning the parameters $\omega_s$ and $\varphi_s$ in each scale using a new graph convolutional network. In the designed AGW, graph wavelet transform is used for localizing graph convolution by projecting signals in the vertex domain into the spectral domain. \par

The obtained AGW with scale $s$ is defined as follows:

\begin{equation}
\label{eq10}
\begin{aligned}
\widetilde{\mathbf{X}}^{(k)}_s\left({\mathbf{:,}}j\right)=\sigma & \Bigg(\mathbf{\Psi }_s\sum^{f_{k-1}}_i{\mathbf{W}^{\left(k\right)}_s(i,j)\mathbf{\Psi }^{-1}_s}{\widetilde{{\mathbf{X}}}}^{\left(k-1\right)}_s(:,i)+\\
& {{\mathbf{V}}}^{\left(k\right)}(i,j){\ }\mathbf{X}^{\left(0\right)}(:,i)\Bigg),\\
& j=1,\dots ,f_k ,
\end{aligned}
\end{equation}

\noindent where $\mathbf{\Psi}_s$ is the wavelet bases at scale $s$ and $\mathbf{\Psi}_s^{-1}$ is its inverse, $\mathbf{W}^{\left(k\right)}_s(i,j)$ and $\mathbf{V}^{\left(k\right)}_s(i,j)$ are learning parameters, $\widetilde{\mathbf{X}}^{(k-1)}_s \in \mathbb{R}^{N\times {f_{k-1}}}$ is the input signal in scale $s$ including $f_{k-1}$ features, $\widetilde{\mathbf{X}}^{(k)}_s \in \mathbb{R}^{N\times {f_{k}}}$ is the output signal including $f_k$ features, $\mathbf{X}^{(0)}$ is the initial node features, and $\sigma(\cdot)$ is the nonlinear activation function.\par

The output of each AGW stack is computed with the average of the outputs of all $\mid\mathcal{S}\mid$ AGWs as $\mathbf{X}^{(k)}=\frac{1}{\mid\mathcal{S}\mid}\sum_{s=1}^{\mid\mathcal{S}\mid} \widetilde{\mathbf{X}}^{(k)}_s $.

\subsection{Multimodal graph wavelet convolution layers}
\label{sec4.2.MultimodalGraphWaveletConvolutionLayers}

The proposed multimodal graph wavelet convolutional network consists of three phases, intra-modality localization, cross-modality correlation, and node classification. Fig. \ref{fig2} schematically shows the diagram of the proposed M-GWCN. The details are given below.

\begin{figure*}[!t]
\centering
\includegraphics[width=\textwidth]{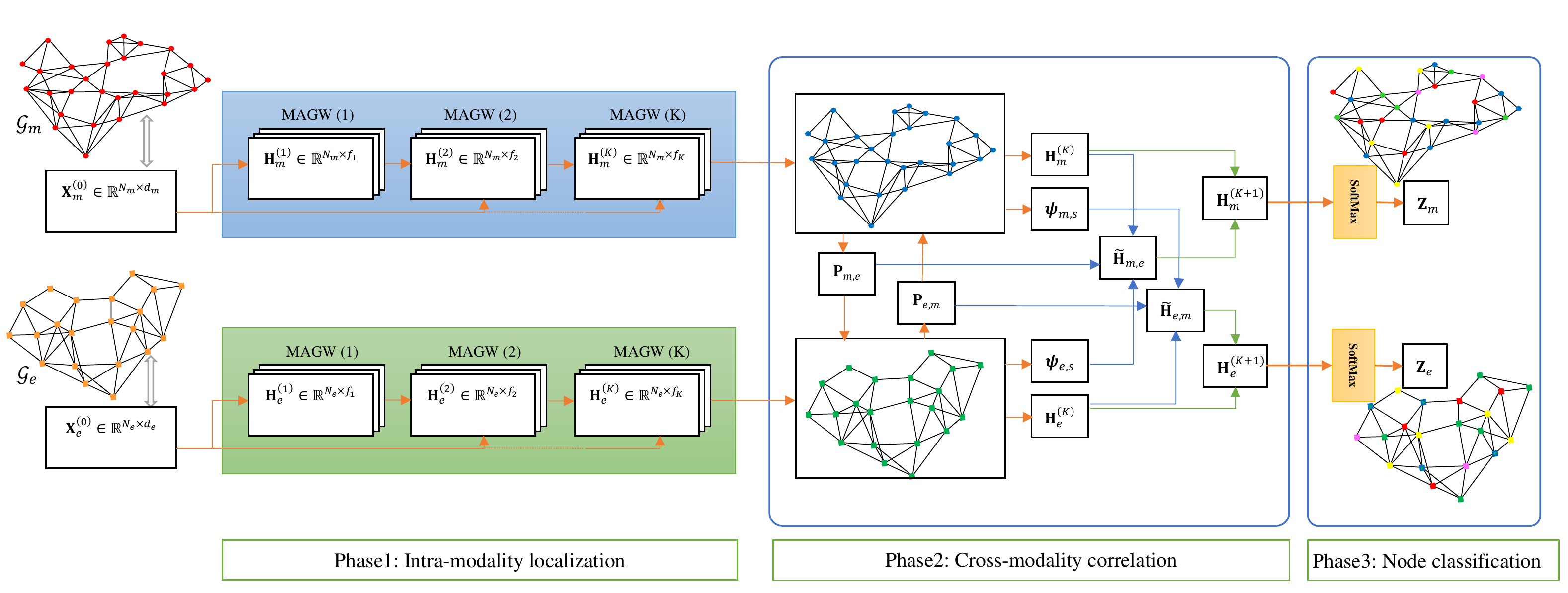}
\caption{Diagram of the proposed M-GWCN model. This model consists of three phases. The first phase consists of the first $K$ layers of the M-GWCN model. Each layer applies multi-scaled graph wavelet convolution for intra-modality localization. In the second phase, a new convolutional layer is defined to explore the cross-modality correlations among various modalities by embedding feature vectors of each modality based on the graph wavelet of the other modalities. Finally, node classification is conducted in the third phase.}
\label{fig2}
\end{figure*}

\subsection*{1. Intra-modality localization}
\label{sec4.2.1.IntraModalityLocalization}

In this phase, $M$ feature mapping processes for node feature vectors of all modalities are conducted separately. For each modality $m$, node features $\mathbf{X}_m\in \mathbb{R}^{N_m\times d_m}$ are represented through $K$ layers of network while each layer contains an $\mid\mathcal{S}\mid$ parallel units of AGWs, where $\mathcal{S}$ is the set of scales. \par

The output of the proposed network for modality $m$ in scale $s$ of $k$-th layer is defined as:

\begin{equation}
\label{eq11}
\widetilde{\mathbf{H}}^{(k+1)}_{m,s}=\sigma \left(\mathbf{\Psi }_{m,s}\uptheta_s\mathbf{\Psi }^{-1}_{m,s}{\widetilde{{\mathbf{H}}}}^{\left(k\right)}_{m,s} \mathbf{W}^{\left(k\right)}_{m,s}+\mathbf{X}^{\left(0\right)}_{m}\mathbf{V}^{\left(k\right)}_{m}\right),
\end{equation}

\noindent where $\mathbf{W}^{\left(k\right)}_{m,s}\in \mathbb{R}^{f_k\times f_{k+1}}$ and $\mathbf{V}^{\left(k\right)}_{m} \in \mathbb{R}^{d_m\times f_{k+1}}$ are trainable parameter matrices for feature mapping in scale $s$, $\uptheta_s$ is a diagonal matrix for graph convolution kernel, $\mathbf{\widetilde{H}}_{m,s}^{(k)}=\left[ \mathbf{\widetilde{H}}_{m,s}^{(k)} {^T}(1,:)...\mathbf{\widetilde{H}}_{m,s}^{(k)}{^T} (N_m,:)\right]^T\in \mathbb{R}^{N_m\times f_k}$ is the embedded feature vectors in layer $k$ and scale $s$, $\mathbf{X}_m^{(0)}$ is the initial node feature in modality $m$, $f_k$ is the number of features in layer $k$, and $\sigma(\cdot)$ is the non-linear activation function. \par

Applying stochastic dropout to the initial node feature in equation (\ref{eq11}) encourages each AGW to provide a response different from the others. \par

The final embedded feature vectors of layer k are defined by averaging the outputs of all $\mid\mathcal{S}\mid$ units of AGW in $k$-th AGW as:

\begin{equation}
\label{eq12}
\mathbf{H}_m^{(k)}=\frac{1}{\mid\mathcal{S}\mid}\sum_{s=1}^{\mid\mathcal{S}\mid} \mathbf{\widetilde{H}}^{(k)}_{m,s}. 
\end{equation}

\subsection*{2. Cross-modality correlations}
\label{sec4.2.2.crossModalityCorrelations}

Tuning the cross-modality correlations or finding the point-wise correspondences among various modalities is the most critical challenge in the multimodal problems. \par
In this phase, the cross-modality correlations on various modalities are explored and a new convolutional layer is defined to represent embedded features of each modality based on the graph wavelet of the other modalities.\par
To prevent the increase of learnable parameters, in this layer, we leverage wavelet with one scale. Our graph wavelet convolutional network learned in phase 1 is permutation invariant because the embedded feature vectors are insensitive to re-ordering the node index. We utilize this property to take advantage of applying correlated representational information of the other modalities discovered by cross-modality correlations. \par

To have better representation of embedded feature vectors in modality $m$ obtained after $K$ layers, $\mathbf{H}_m^{(K)}$, based on the wavelet bases of modality $e$ $(e\neq m)$, we define the following cross-modality feature mapping between two modalities $m$ and $e$:

\begin{equation}
\label{eq13}
\widetilde{\mathbf{H}}_{m,e}=\mathbf{P}_{m,e} \left( \mathbf{\Psi }_{e,s} \uptheta_{e,s}  \mathbf{\Psi }^{-1}_{e,s} \right) \mathbf{P}_{m,e}^T \mathbf{H}_m^{(K)},
\end{equation}

\noindent where $\mathbf{P}_{m,e}$ is a $N_m\times N_e$ permutation matrix that encoded cross-modality correspondence between modalities $e$ and $m$, and $\widetilde{\mathbf{H}}_{m,e}\in \mathbb{R}^{N_m\times f_K}$ is a matrix of embedded feature vectors in modality $m$ obtained by representing $\mathbf{H}_m^{(K)}$ based on the wavelet bases of modality $e$, while their correlations are encoded in $\mathbf{P}_{m,e}$. \par

A permutation matrix is defined as a matrix including exactly one single unit value in each row and column, and zeros elsewhere. This matrix is used to represent the permutations of elements in an ordered sequence. \par

For example, for the following square matrix $\mathbf{A}$ with $3$ rows, a permutation in order of its rows as $(3,1,2)$ is represented by the shown permutation matrix $\mathbf{P}$, which yields $\mathbf{A}_P=\mathbf{P}\mathbf{A}$ by a simple matrix-vector multiplication, as:

\begin{align*}
\mathbf{A}=&\begin{bmatrix}a_{1,1} & a_{1,2} & a_{1,3} \\a_{2,1} & a_{2,2} & a_{2,3} \\ a_{3,1} & a_{3,2} & a_{3,3}\end{bmatrix}, \mathbf{P}=\begin{bmatrix}0 &0 &1 \\1 & 0 &0 \\ 0 & 1 & 0\end{bmatrix},\\
 &\mathbf{A}_P=\begin{bmatrix} a_{3,1} & a_{3,2} & a_{3,3} \\a_{1,1} & a_{1,2} & a_{1,3} \\ a_{2,1} & a_{2,2} & a_{2,3}\end{bmatrix}.
\end{align*}

The permutation of symmetric matrix $\mathbf{A}$, in order of both rows and columns, is obtained by matrix-vector multiplication $\mathbf{A}_P=\mathbf{PAP}^T$.\par

Based on the cross-modality feature mapping, for each modality $m$, we conduct a cross-modality convolutional layer as follows:

\begin{equation}
\label{eq14}
\mathbf{H}_{m}^{(K+1)}=\sigma \left( con \left( \mathbf{H}_{m}^{(K)},\widehat{\mathbf{H}}_{m} \right) \mathbf{W}_m^{(K)} \right),
\end{equation}

\noindent where $\mathbf{H}_{m}^{(K+1)} \in \mathbb{R}^{N\times f_{K+1}}$ is the output embedded feature matrix, $\mathbf{W}_m^{(K)}\in R^{(Mf_K)\times f_{K+1}}$ is kernel matrix, $\widehat{\mathbf{H}}_m=con\left(\widetilde{\mathbf{H}}_{m,1},...,\widetilde{\mathbf{H}}_{m,m-1},\widetilde{\mathbf{H}}_{m,m+1},...,\widetilde{\mathbf{H}}_{m,M}\right)$, $K$ is the total number of layers for intra-modality representation, and $con(\cdot)$ function is the concatenation operator that incorporates the extracted feature information in the cross-modality convolution layer.

\subsection*{3. Node classification}
\label{sec4.2.3.nodeClassification}

The embedded feature vectors of each modality are obtained through $K$ layers of feature mapping (intra-modality representation) and one layer of feature mapping based on the other modalities (cross-modality representation). \par

The obtained embedded feature vectors are fed into the last layer to conduct node classification as follows:

\begin{equation}
\label{eq15}
\mathbf{Z}_{m}=softmax \left( \mathbf{H}_{m}^{(K+1)}\mathbf{W}_m^{(K+1)} \right),
\end{equation}

\noindent where $\mathbf{W}_m^{(K+1)}\in \mathbb{R}^{f_{K+1}\times C}$ , $\mathbf{Z}_{m} \in \mathbb{R}^{N_m\times C}$, and $C$ is number of classes.

\subsection*{4. Problem formulation and optimization }
\label{sec4.2.4.problemFormulationAndoptimization}

Considering the architecture of the proposed network, we present a unified regularization-based optimization problem that aims to simultaneously learn network parameters and permutations through network training.\par

Since the permutations are highly discrete and too costly to enumerate, the stochastic gradient descent (SGD) method is not capable to optimize the proposed networks because it applies for optimizing networks with continuous parameters.\par

We extend our formulation to the nearest convex surrogate by approximating all permutation matrices $\mathbf{P}_{m,e} (1\leq m,e\leq M, {\ } m\neq e)$ in equation (\ref{eq13}) with doubly stochastic matrices $\widetilde{\mathbf{P}}_{m,e}$. A doubly stochastic matrix $\widetilde{\mathbf{P}}_{m,e}$ is an $N_m\times N_e$ matrix of non-negative real numbers, each of whose rows and columns sums to $1$, i.e.:

\begin{align*}
\widetilde{p}_{m,e}(i,j)\ge 0,{\ } \widetilde{\mathbf{P}}_{m,e} \mathbf{1}_e=\mathbf{1}_m, {\ } \widetilde{\mathbf{P}}_{m,e}^T \mathbf{1}_m=\mathbf{1}_e,
\end{align*}

\noindent where $\mathbf{1}_e$ is an $N_e$-dimensional column vector of ones.\par

According to this relaxation, we propose a new network loss function as:

\begin{equation}
\label{eq16}
\mathcal{L}=\mathcal{L}_{CE}(\mathcal{W},\mathcal{P})+\alpha \mathcal{L}_{DSM} (\mathcal{P}), 
\end{equation}

\noindent where $\mathcal{W}=\left\{ \mathbf{W}_1^{(k)},...,\mathbf{W}_M^{(k)} \right\}_{k=1}^K$ is the set of network weights, $\mathcal{P}=\left\{\widetilde{\mathbf{P}}_{m,e} \mid 1\leq m,e\leq M, {\ } m \neq e \right\}$ is the set of doubly stochastic matrices used in extension of equation (\ref{eq13}), $\mathcal{L}_{CE} (\mathcal{W},\mathcal{P})$ is the categorical cross-entropy loss function, that will be computed by equation (\ref{eq18}), $\mathcal{L}_{DSM} (\mathcal{P})$ is the loss function for doubly stochastic matrix, which will be given in equation (\ref{eq19}), and $\alpha$ is a trade-off parameter between them.\par

$\mathcal{L}_{CE} (\mathcal{W},\mathcal{P})$ measures the empirical loss on the training data by summing up the discrepancy between the outputs of the network $\mathbf{Z}_m$ and the ground-truth $\mathbf{Y}_m$ for all modalities $m=1,...,M$:

\begin{equation}
\label{eq17}
\mathcal{L}_{CE}(\mathcal{W},\mathcal{P})=\sum_{m=1}^M \sum_{i=1}^{l_m} \mathbf{y}_{m,i} {\ } ln \mathbf{Z}_m (i,:),
\end{equation}

\noindent where $\mathbf{Z}_m$ is the output of the network, obtained in equation (\ref{eq15}). This output is a function of the set of network weights $\mathcal{W}$ in equations (\ref{eq11}), (\ref{eq14}), and (\ref{eq15}) and the set of doubly stochastic matrices $\mathcal{P}$ in the extension of equation (\ref{eq13}).\par

The loss function for learning the doubly stochastic matrix $\widetilde{\mathbf{P}}_{m,e}$ between two modalities $m$ and $e$ is as follows:

\begin{equation}
\label{eq18}
\begin{aligned}
\mathcal{L}_{DSM}(\widetilde{\mathbf{P}}_{m,e})=&\sum_{i=1}^{N_m} \left( \sum_{j=1}^{N_e} \mid \widetilde{p}_{m,e}(i,j)\mid -1 \right) +\\
 &\sum_{j=1}^{N_e} \left( \sum_{i=1}^{N_m} \mid \widetilde{p}_{m,e}(i,j)\mid -1 \right).
\end{aligned}
\end{equation}

Thus, the loss function considering all doubly stochastic matrices is defined as follow:

\begin{equation}
\label{eq19}
\mathcal{L}_{DSM}(\mathcal{P})=\sum_{m=1}^M \sum_{e=1,\\ e<m}^M \mathcal{L}_{DSM}(\widetilde{\mathbf{P}}_{m,e}).
\end{equation}

The final optimization problem is also included the following additional regularization terms:

\begin{equation}
\label{eq20}
\mathcal{R}=\beta \sum_{m=1}^M \sum_{k=1}^{K+1} \parallel \mathbf{W}_m^{(k)}\parallel_F^2+\gamma \mathcal{R}_{BM}(\mathcal{P})+\lambda \mathcal{R}_{WM}(\mathcal{W}),
\end{equation}

\noindent where $\beta$, $\gamma$, and $\lambda$ are regularization parameters. The first term is the $L2$-norm regularization penalty on the network weights preventing overfitting by constraining the complexity of the learned kernels of all layers. The second term is the between-modality regularization term that restricts the output labels of all modality pairs m and e based on cross-modality relations encoded in permutation matrix $\widetilde{\mathbf{P}}_{m,e}$ as follows:

\begin{equation}
\label{eq21}
\begin{aligned}
\mathcal{R}_{BM}(\mathcal{P})= \sum_{m=1}^M \sum_{e=1,m<e}^{M} & \Big( \parallel \widetilde{\mathbf{P}}_{m,e}^T \mathbf{Z}_m -\mathbf{Z}_e \parallel_F^2+ \\
&\parallel \mathbf{Z}_m - \widetilde{\mathbf{P}}_{m,e} \mathbf{Z}_e \parallel_F^2 \Big).
\end{aligned}
\end{equation}

The third term of equation (\ref{eq20}) is the intra-modality regularization term, which is defined based on preserving manifold constraints of multimodal data based on the following equation:

\begin{equation}
\label{eq22}
\mathcal{R}_{WM}(\mathcal{W})= \sum_{m=1}^M \sum_{i,j=1 , i<j}^{N_m} a_m^{(i,j)} \parallel \mathbf{Z}_{m}(i,:)-\mathbf{Z}_m(j,:) \parallel^2,
\end{equation}

\noindent where $a_m^{(i,j)}$ is the weight of edge between $i$-th and $j$-th vertices. \par

According to loss functions and regularization terms defined in equations (\ref{eq16}) and (\ref{eq20}), the final optimization problem on the objective function $\mathcal{J}=\mathcal{L}+\mathcal{R}$ is obtained as follows:

\begin{equation}
\label{eq23}
\begin{aligned}
\mathcal{J}=  \min_{\mathcal{W},\mathcal{P}} {\ } & \mathcal{L}_{CE}(\mathcal{W},\mathcal{P})+ \alpha  \sum_{m=1}^M \sum_{e=1, m<e}^M \mathcal{L}_{DSM}(\widetilde{\mathbf{P}}_{m,e})+\\
&\beta  \sum_{m=1}^M \sum_{k=1}^{K+1}\parallel \mathbf{W}_m^{(k)}\parallel_F^2+\gamma \mathcal{R}_{BM}(\mathcal{P})+\lambda \mathcal{R}_{WM}(\mathcal{W}),\\
&s.t. {\ } \widetilde{p}_{m,e}(i,j)>0,{\ } 1\leq m,e\leq M.
\end{aligned}
\end{equation}

To optimize this problem under its constraints, an iterative optimization algorithm can be adopted to alternatively minimize loss function with respect to $\mathcal{W}$ and $\mathcal{P}$.\par

In the first pass, $\nabla_{\mathbf{W}_m^{(k)}} \mathcal{J}$, that is the gradient of $\mathcal{J}$ with respect to $\mathbf{W}_m^{(k)}$ is computed while all other parameters are considered fixed. According to the stochastic gradient descent optimization method, all network kernel matrices $\mathbf{W}_m^{(k)} (m=1,...,M and {\ } k=1,...,K+1)$ can be update using the following iterative equation until convergence:

\begin{equation}
\label{eq24}
\mathbf{W}_m^{(k)}(t+1)=\mathbf{W}_m^{(k)}(t)-\eta \nabla_{\mathbf{W}_m^{(k)}} \mathcal{J} (t+1),
\end{equation}

\noindent where $\eta$ is the learning rate of the SGD method and $t$ is iteration number.\par

At the second pass, $\nabla_{\widetilde{\mathbf{P}}_{m,e}} \mathcal{J}$ is computed as the gradient of $\mathcal{J}$ with respect to $\widetilde{\mathbf{P}}_{m,e}{\ }(1\leq m,e\leq M, {\ } m\neq p)$ while all other parameters are fixed. Permutation matrices $\widetilde{\mathbf{P}}_{m,e}$ can be updated using SGD as:

\begin{equation}
\label{eq25}
\widetilde{\mathbf{P}}_{m,e} (t+1)=\widetilde{\mathbf{P}}_{m,e} (t)-\eta \nabla_{\widetilde{\mathbf{P}}_{m,e}} \mathcal{J} (t+1),
\end{equation}

\noindent while the non-negative constraint is maintained by thresholding $\widetilde{\mathbf{P}}_{m,e}=max(\widetilde{\mathbf{P}}_{m,e},0)$. \par

Practically, for optimizing the above-mentioned problems, we take advantage of Keras \footnote{\url{https://keras.io/api/losses/}} library by adding a new loss function and defining regularization terms on each layer, which these penalties are summed into the loss function during optimization.\par

The proposed M-GWCN is summarized in Algorithm \ref{alg1}.

\begin{algorithm}[h]
\caption{Multimodal Graph Wavelet Convolutional Network (M-GWCN).}
\label{alg1}
\begin{algorithmic}
\STATE
\STATE {\textsc{Inputs}}
\begin{itemize} 
    \item $M$: number of modalities
    \item $\mathcal{G}_m=(\mathcal{V}_m,\mathcal{E}_m,\mathbf{A}_m)$: Undirected weighted graph of modality $m$ $(1\leq m\leq M)$
    \item $\mathbf{Y}_m=\left[\mathbf{y}_{m,1}...\mathbf{y}_{m,{N_m}} \right]^T$: Label vector of data samples in modality $m$ $(1\leq m\leq M)$
\end{itemize}
 
\STATE {\textsc{Hyper-parameters}}
\begin{itemize}
    \item $\mathcal{S}$: Set of scales
    \item $K$: Number of layers
    \item $f_k$: Number of features in layer $k$
    \item $r$: Dropout rate for initial node feature
    \item $\sigma(\cdot)$: Nonlinearity activation function
    \item $Q$: Number of Chebyshev polynomials
    \item $\alpha$, $\beta$, $\gamma$, $\lambda$ : Regularization parameters
    \item $T$: Maximum iteration number
\end{itemize}

\STATE {\textsc{Steps}}
\begin{enumerate}
    \item Approximate $\mathbf{\Psi}_{m,s}$ for each modality $m$ in scale $s$ using equation (\ref{eq8}).
    \item Initialize $\mathbf{W}_m^{(k)}(0)$ and $\widetilde{\mathbf{P}}_{m,e} (0)$ randomly for all modalities $m,e=1,...,M$ and all layers $k=1,...,K+1$.
    \item \textbf{For} ($t=1$ to $T$) \textbf{do}:
    \begin{enumerate}
        \item Compute $\nabla_{\mathbf{W}_m^{(k)}} \mathcal{J}$ and update kernel matrix $\mathbf{W}_m^{(k)}$ using equation (\ref{eq24}).
        \item Compute $\nabla_{\widetilde{\mathbf{P}}_{m,e}} \mathcal{J}$ and update doubly stochastic matrices $\widetilde{\mathbf{P}}_{m,e}$ using equation (\ref{eq25}).
        \item Maintain non-negative constraint by thresholding $\widetilde{\mathbf{P}}_{m,e}=max(\widetilde{\mathbf{P}}_{m,e},0)$.
    \end{enumerate}
    \textbf{End for} 
\end{enumerate}

\STATE {\textsc{Outputs}}
\begin{itemize}
    \item Network parameters $\mathbf{W}_m^{(k)} (m=1,...,M {\ } and {\ } k=1,...,K+1)$.
    \item 	Doubly stochastic matrices $\widetilde{\mathbf{P}}_{m,e} {\ } (1\leq m,e \leq M)$.
\end{itemize}

\end{algorithmic}
\end{algorithm}

\subsection*{5. Computational complexity analysis}
\label{sec4.2.5.computationalComplexityAnalysis}

The computations of the M-GWCN consists of two parts, computing graph wavelet bases and computing the network outputs.\par

Since M-GWCN employs the Chebyshev polynomials in approximating graph wavelet bases, it takes its linearity advantage with computational complexity $O(E_m\times Q)$, where $E_m$ is the number of edges of $\mathcal{G}_m$ and $Q$ is the order of Chebyshev polynomials. \par

Under the assumption of sparse wavelet bases, each scale s in modality $m$ can be implemented as a matrix multiplication between sparse square matrix $\mathbf{\Psi}_{m,s} \uptheta_s \mathbf{\Psi}_{m,s}^{-1}$ and embedded feature matrix $\mathbf{H}_{m,s}^{(k)}$, which has incurred linear complexity for each layer $k$ in feature mapping phase. \par

The computation complexity in layer $K+1$ of each modality depends on computing cross-modality correlation. According to the sparsity nature of both permutation matrices and wavelet bases, computing the cross-modality correlation using matrix multiplication also has linear complexity.\par

Finally, according to the linear time of matrix multiplication between embedded feature matrix $\mathbf{H}_m^{(K+1)}$ and kernel weight matrix $\mathbf{W}_m^{(K+1)}$ in each modality $m$, and also the linearity of computing sigmoid function, the label of each layer in the final layer (classification) can be estimated in linear time.

\subsection{Other versions of M-GWCN}
\label{4.3.otherVersionsOfM-GWCN}

\subsection*{1. Graph wavelet convolutional Network (GWCN)}
\label{sec4.3.1.graphWaveletConvolutionalNetwork}

Since most of the explicit graph-based data are unimodal, we simplify the proposed M-GWCN for unimodal tasks. GWCN, as a unimodal version of M-GWCN, is designed using the first $K$ layers integrated with the last classification layer (without cross-modality correlations).

\subsection*{2. Multi-view graph wavelet convolutional Network (MV-GWCN)}
\label{sec4.3.2.multi-viewGraphWaveletConvolutionalNetwork}

Many multimodal problems get the benefits from the prior knowledge of fully or partially correspondence information among modalities. Since these problems, sometimes called multi-view, are special cases of multimodal problems, we redesigned our proposed M-GWCN to cope with multi-view data, called MV-GWCN. In MV-GWCN, permutation matrices encode correspondence information among various modalities, such that if sample $i$ in modality $m$, $\mathbf{x}_{m,i}$, corresponds with sample $j$ in modality $e$, $\mathbf{x}_{e,j}$, then $p_{m,e} (i,:)$ contains $1$ in $j$-th entry and $0$ otherwise. \par
MV-GWCN is trained similar to M-GWCN, while permutation matrices are considered as non-learning parameters.

\section{Experiments}
\label{sec5.experiments}

We investigate the effectiveness of the proposed network with two types of experiments on unimodal explicit graph-based data and multimodal implicit graph-based ones.\par

The first experiment examines the efficiency of the proposed network on inherently graph-based data, including citation datasets, considering semi-supervised node classification tasks. The purpose of second experiment is to evaluate the effectiveness of the proposed method on multimodal data, implicitly considered as a graph, compared with state-of-the-art semi-supervised multimodal problems.
Public implementations with the open-source GNN libraries Spektral \cite{Wang2020a} (TensorFlow/Keras) are available in \url{https://github.com/maysambehmanesh/MGWCN}.

\subsection{Evaluation on unimodal explicit graph-based data}
\label{sec5.1.evaluationOnUnimodalExplicitGraph-basedData}

To evaluate the performance of M-GWCN on explicit graph-based data, we focus on semi-supervised node classification of popular citation datasets. Since these datasets are unimodal, we apply the unimodal version of M-GWCN (GWCN) for semi-supervised node classification. \par

Specifications of three benchmark datasets used in the experiments, including the number of nodes, edges, node features, and classes are reported in Table \ref{table1}.

\begin{table}[!t]
\caption{The properties of citation datasets}
\label{table1}
\centering
\begin{tabular}{lcccc}
\hline
\textbf{Dataset}  & \textbf{Nodes} & \textbf{Edges} & \textbf{Features} & \textbf{Classes} \\ \hline
Cora     & 2708           & 5429           & 1433              & 7                \\
Citeseer & 3327           & 9228           & 3703              & 6                \\
Pubmed   & 19717          & 88648          & 500               & 3                \\ \hline
\end{tabular}
\end{table}

We compare the efficiency of GWCN with most popular state-of-the-art graph convolutional networks. These baselines include ChebyNet \cite{Defferrard2016}, GCN \cite{Kipf2017}, CayleyNets \cite{Levie2019}, GNN-ARMA \cite{Bianchi2021}, and GWNN \cite{Xu2019} as spectral methods, in addition to GAT \cite{Velickovic2018}, GraphSAGE \cite{Hamilton2017}, and GIN \cite{Xu2019a} as spatial methods. \par

In the semi-supervised problem, the features of all vertices are known, but only $20$ labels per class are given for training. Also, $500$ and $1000$ labeled nodes are considered for validation and test, respectively.\par

The task is learning a network that takes the feature vectors as inputs and assigns a label to each vertex as outputs. The evaluation results of the learned network on testing nodes are reported as classification accuracies.\par

Table \ref{table2} reports experimental results of GWCN on citation data sets described in Table \ref{table1}, and compares them with state-of-the-art methods. The accuracies of CayleyNets, GraphSAGE, and GIN have been reported from the respective paper while others implemented by authors according to mentioned settings. In this table, GWCN-1 indicates GWCN model with only one and the first scale, mentioned in Table \ref{table3}, e.g. $0.7$ for Cora, and GWCN-2 considers both scales.\par

The maximum number of training epochs is set to $20000$. Training of network will be terminated if the validation loss does not decrease for $50$ consecutive epochs. Learning rate is $\eta=0.01$ for all methods.\par

For maintaining the sparsity structure of graph wavelet bases, a threshold $t$ is define to refine the values of  $\mathbf{\Psi}_{m,s}$ and $\mathbf{\Psi}_{m,s}^{(-1)}$ such that the value of entries that are smaller than threshold $t$ is set to $0$. This parameter in all experiments is set to $t=10^{-4}$.

\begin{table}[!t]
\caption{Accuracies of different methods on citation datasets (Mean $\pm$ Standard deviation)}
\label{table2}
\centering
\begin{tabular}{lccc}
\hline
\textbf{Method}     & \textbf{Cora}                & \textbf{Citeseer} & \textbf{Pubmed} \\ \hline
GIN        & 75.1±1.7                     & 63.1±0.2          & 77.1±0.7        \\
GraphSAGE  & 73.7±1.8                     & 65.9±0.9          & 78.5±0.6        \\
GAT        & 83.1±0.4                     & 70.7±0.1          & 78.4±0.3        \\
ChebyNet   & 78.2±0.6                     & 68.6±0.2          & 76.3±0.8        \\
GCN        & 82.3±0.4                     & 71.4±0.3          & 79.3±0.2        \\
CayleyNets & \multicolumn{1}{r}{81.2±1.2} & 67.1±2.4          & 75.6±3.6        \\
GNN-ARMA   & 81.4±0.4                     & 68.9±0.6          & 76.3±0.3        \\
GWNN       & 82.6±0.3                     & 70.4±0.9          & 79.4±0.7        \\
GWCN-1     & 83.2±0.1                     & 71.6±0.6          & 79.5±0.3        \\
GWCN-2     & \textbf{83.8±0.5}            & \textbf{72.7±0.3} & \textbf{80.3±0.6} \\ \hline
\end{tabular}
\end{table}

As proved in \cite{Hammond2011}, using small values for scale parameters, graph wavelet bases induce locality properties in such a way that each basis represents the neighborhood structure of a specific vertex. To avoid model complexity in this experiment, we use two scales for graph wavelets, $\mid\mathcal{S}\mid=2$. The values of these scales are chosen among small values between $0$ and $1$ via grid search to ensure the locality of convolution in the vertex domain. The values of all parameters of GWCN used in the above experiments are summarized in the Table \ref{table3}. Parameters of other methods are chosen according to their respected papers.

\begin{table}[!t]
\caption{Parameters of the GWCN model}
\label{table3}
\centering
\begin{tabular}{lccc}
\hline
\textbf{Parameter}    & \textbf{Cora} & \textbf{Citeseer} & \textbf{Pubmed} \\ \hline
$K$          & 2             & 2                 & 1               \\
$\mathcal{S}$         & $\left\{0.7,1\right\}$    & $\left\{0.5,0.7\right\}$  & $\left\{0.3,0.5\right\}$                \\
$t$                   & $10^{-4}$           & $10^{-5}$             & $10^{-7}$        \\
$Q$                   & 40                  & 40                    & 30               \\
$r$                   & $0.75$              & $0.75$                & $0.25$           \\
$\beta$              & $5\times 10^{-4}$   & $5\times 10^{-4}$     & $5\times 10^{-4}$ \\ \hline
\end{tabular}
\end{table}

According to classification accuracies reported in Table \ref{table2}, GWCN provides the best accuracies among all baselines on all datasets. The second-best accuracies for Cora and Pubmed are provided by GWNN and for Citeseer are reported with GCN. These results demonstrate the capability of graph wavelet bases in achieving a better representation of vertex domain compared to other spectral methods. \par

Furthermore, GWCN performs better than all spatial methods reflecting the promising ability of spectral methods to achieve good performance.  Since GAT assigns a self-attention weight to each edge, it captures more effectively the local similarity among neighborhoods and provides better accuracy compared to other spatial methods. However, computing the attention weights is inefficient for the large number of edges, and unlike GWCN, GAT cannot take advantage of the global structure of graphs. \par

Among spectral methods, GCN not only consistently outperforms others but also achieves a significant accuracy compared with GWCN. \par

The key to the good performance of GCN is that, unlike CayleyNets that applies graph Fourier bases to express the spectral graph convolution, GCN leverages the Laplacian matrix as a weighted matrix in its formulation (equation (\ref{eq5})), which in terms of sparsity, Laplacian matrix is sparser than Fourier bases and is similar to wavelet bases. \par

GWNN presents closer results to GWCN because it employs wavelet bases to localized graph convolution. Nevertheless, GWCN has two main advantages that enable it to consistently outperforms GWNN: 1) Aggregating features with different scaling parameter values makes GWCN flexible in locally exploring each sub-graph according to an appropriate scale, instead of using only one scale parameter for the whole graph. 2) Due to the sparsity of wavelet bases, after a few convolutions, the node features in GWNN become too smooth. Since GWCN formulation is adopted with a residual connection, it can be amplified with the initial node features avoiding over-smoothing.\par

Table \ref{table4} reports the training time and parameter complexity of GNN methods.

\begin{table}[!t]
\caption{Training time and parameter complexity of GNN methods}
\label{table4}
\centering
\begin{tabular}{lcccc}
\hline
\multirow{2}{*}{\textbf{Method}} & \multicolumn{2}{c}{\textbf{Cora}} & \multicolumn{2}{c}{\textbf{Citeseer}} \\ \cline{2-5}
                                 & Sec/epoch       & Parameters      & Sec/epoch         & Parameters        \\ \hline
ChebyNet                  & 0.28492         & 46080           & 0.69860           & 118688         \\
GAT                       & 0.26980         & 92373           & 0.55968           & 237586         \\
GCN                       & 0.23140         & 23040           & 0.56030           & 59344          \\
GNN-ARMA                  & 0.23029         & 46103           & 0.37047           & 118710         \\
GWNN                      & 0.34075         & 23063           & 0.46709           & 59366          \\
GWCN-1                    & 0.37979         & 46103           & 0.54539           & 178054         \\
GWCN-2                    & 0.47666         & 91975           & 0.61975           & 355814         \\ \hline
\end{tabular}
\end{table}

Since in citation network, Laplacian is sparser than graph wavelets \cite{Xu2019}, two graph wavelet-based methods, GWNN and GWCN, have a little more time complexity compared with Laplacian-based methods, ChebyNet, GCN, and GNN-ARMA. \par

According to the definition of GNN models, most of them have moderate number of parameters because increasing their parameters by adding multiple layers will raise the over-smoothing problem. But, due to two main reasons, increasing the parameters of GWCN model does not lead to over-smoothing: 1) these additional parameters are respected to multiple stacks in each layer with different scaling parameter values, and not respected to more layers, 2) residual connection adopted for GWCN prevents the over-smoothing of each stack. Also, due to effective computations provided by sparse wavelet bases, GWCN achieves better performance in reasonable time complexity. Therefore, unlike most GNN models, the performance of GWCN is improved using additional parameters. \par

\subsection{Evaluation of multimodal implicit graph-based data}
\label{sec5.2.evaluationOfMultimodalImplicitGraph-basedData}

The primary purpose of the proposed method is generalizing graph convolutional neural networks for multimodal problems. This experiment evaluates the effectiveness of the M-GWCN method for multimodal data. \par

To evaluate the performance of the proposed method, we conduct M-GWCN for the classification of multimodal implicit graph-based data on two categories of benchmark problems, including multimodal and multi-view datasets.\par

These categories consist of two multimodal datasets, Caltech and NUS, and three multi-view datasets, Caltech101-7, Caltech101-20, and MNIST, as introduced in \cite{Behmanesh2021}. In these datasets, each modality is implicitly defined as a graph. \par

In the first experiment, we evaluate M-GWCN on multimodal datasets in a more practical scenario, without any predefined knowledge among modalities. To ensure this property, we randomly shuffle the order of data samples in each modality, which makes the point-wise correspondences between the original modalities unknown.\par

We compare the efficiency of M-GWCN with most state-of-the-art multimodal data modeling methods including CD (pos) \cite{Eynard2015}, CD (pos+neg) \cite{Eynard2015}, SCSMM \cite{Pournemat2021}, m-LSJD \cite{Behmanesh2021}, m$^2$-LSJD \cite{Behmanesh2021}, and M$^2$CPC-u \cite{Behmanesh2021CrossModalAM}.\par

The first experiment conducts M-GWCN on multimodal datasets with the parameters in Table \ref{table5}. In this experiment, data in each modality is portioned to $50\%$ for training, $30\%$ for validation, and $20\%$ for testing. The maximum number of training epochs is $20000$, and the training phase will be terminated if the validation loss does not decrease for $100$ consecutive epochs. Experimental results over $10$ randomly data splits are repeated in terms of mean accuracy and standard deviations. Parameters of other methods are chosen according to their respective paper.

\begin{table}[!t]
\caption{Parameters of M-GWCN model}
\label{table5}
\centering
\begin{tabular}{lcc}
\hline
\textbf{Parameter} & \multicolumn{1}{c}{\textbf{Caltech}} & \multicolumn{1}{c}{\textbf{NUS}} \\ \hline
$K$                  &2                                     &2                                  \\
$\mathcal{S}$        &$\left\{ 0.7,0.9 \right\}$            & $\left\{ 0.5,0.7 \right\}$       \\                         $t$                  &$10^{-4}$                             & $10^{-5}$        \\ 
$Q$                  &$50$                                  & $40$        \\ 
$r$                  &$0.75$                                & $0.75$        \\ 
$\eta$               &$10^{-2}$                             & $10^{-2}$        \\ 
$\alpha$             &$10^{-6}$                             & $10^{-6}$        \\ 
$\beta$              &2000                                  & 1000        \\ 
$\gamma$             &100                                   & 100        \\ 
\hline
\end{tabular}
\end{table}

Table \ref{table6} reports the performance of M-GWCN in terms of mean and standard deviation of accuracies, and compares it with baselines. Among these compared methods, M$^2$CPC-u has the most similar experimental settings with M-GWCN because it lakes prior correspondence knowledge. Although other methods have been developed for multimodal datasets, they still are more or less dependent on prior knowledge about correspondences among modalities. In the respective papers of methods CD (pos), CD (pos+neg), m-LSJD, and m$^2$-LSJD, the rate of dependency is defined based on the ratio of the number of given corresponding samples to the total number of samples (in percent). To get closer to a fair comparison, we consider the correspondence ratio as minimum as possible ($10\%$) in all mentioned methods. \par

\begin{table}[!t]
\caption{Classification accuracies on multimodal datasets (Mean $\pm$ Standard Deviation)}
\label{table6}
\centering
\begin{tabular}{lcc}
\hline
\textbf{Method} & \textbf{Caltech}  & \textbf{NUS}      \\ \hline
CD (pos)        & 76.9±1.1          & 81.8±0.8          \\
CD (pos+neg)    & 73.4±0.8          & 80.3±0.4          \\
SCSMM           & -                 & 83.9±2.42         \\
m-LSJD          & 84.1±1.4          & 83.2±1.3          \\
m$^2$-LSJD         & 88.5±1.6          & 87.2±1.1          \\
M$^2$CPC-u         & 84.8±0.7          & 86.4±0.3          \\
M-GWCN          & \textbf{90.6±0.4} & \textbf{89.2±0.8} \\ \hline
\end{tabular}
\end{table}

According to classification accuracies reported in Table \ref{table6}, M-GWCN achieves a significant improvement and consistently outperforms other methods. These results indicate the superiority of the proposed graph convolutional network in effectively finding the cross-modal correlations in absence of given coupling/decoupling information between various modalities.\par

The M-GWCN is able to represent feature vectors of each modality based on the correlated wavelet bases on other modalities. This capability enables it to be not only successfully applicable to the multimodal graph-based data, but also usable efficiently independent of initial correspondence information. \par

Since in many multimodal methods, the correspondences among modalities are predetermined, for more experiments, we evaluate the performance of our proposed network on the multi-view datasets. Therefore, in the second experiment, we use MV-GWCN as a particular version of M-GWCN and conduct it on multi-view graph-based datasets, according to parameters listed in Table \ref{table7}. The other settings are similar to the first experiment. \par

\begin{table}[!t]
\caption{Parameters of MV-GWCN model}
\label{table7}
\centering
\begin{tabular}{lccc}
\hline
\textbf{Parameter}    & \textbf{Caltech101-7} & \textbf{Caltech101-20} & \textbf{MNIST} \\ \hline
$K$          & 2             & 2                 & 2               \\
$\mathcal{S}$         & $\left\{0.5,0.7\right\}$    & $\left\{0.7,0.8\right\}$  & $\left\{0.3,0.6\right\}$                \\
$t$                   & $10^{-6}$           & $10^{-4}$             & $10^{-5}$        \\
$Q$                   & 40                  & 30                    & 35               \\
$r$                   & $0.75$              & $0.75$                & $0.75$           \\
$\eta$                & $10^{-2}$           & $10^{-2}$             & $10^{-2}$ \\ \hline
\end{tabular}
\end{table}

To evaluate the performance of the proposed method on multi-view datasets, we compare the efficiency of MV-GWCN with some existing methods evaluated on mentioned multi-view datasets.  The state-of-the-art methods used for comparison include MLDA \cite{Cao2018}, MLDA-m \cite{Cao2018}, MULDA \cite{Cao2018}, MULDA-m \cite{Cao2018}, MvMDA \cite{Sun2016}, OGMA \cite{Wang2020OrthogonalMA}, OMLDA \cite{Wang2020OrthogonalMA}, OMvMDA \cite{Wang2020OrthogonalMA}, and M$^2$CPC-p \cite{Behmanesh2021CrossModalAM}. \par

Table \ref{table8} reports the classification accuracy obtained by the M-GWCN compared with different multimodal problems on three multi-view datasets. In this table, MV-GWCN-1 indicates the proposed method with only one and the first scale mentioned in Table \ref{table7}. Similarly, MV-GWCN-2 is based on both scales. \par

As can be seen from Table \ref{table8}, MV-GWCN consistently outperforms other methods. According to these results, although our proposed method achieves the best accuracy among all methods on all multi-view datasets, the differences between their accuracies are not as significant as multimodal datasets. The main reason is that on multi-view datasets, the knowledge of correspondence is predetermined and corresponded samples can be fused into a joint latent space for boosting the classification, which improvs accuracy without needing to explore the cross-modality correlations.  When exploring the cross-modality correlations for discovering the correspondences among various modalities is a necessity, because of the lack of this type of prior knowledge, the superiority of our proposed method is proved to be more significant.\par

\begin{table}[!t]
\caption{Classification accuracies on multi-view datasets (Mean $\pm$ Standard Deviation)}
\label{table8}
\centering
\begin{tabular}{lccc}
\hline
\textbf{Method} & \textbf{Caltech101-7} & \textbf{Caltech101-20} & \textbf{MNIST}     \\ \hline
MLDA            & 92.29$\pm$8e-3            & 76.59$\pm$12e-3            & 92.84$\pm$5e-3         \\
MLDA-m          & 89.78$\pm$10e-3           & 73.77$\pm$114e-3           & 93.09$\pm$8e-3         \\
MULDA           & 92.65$\pm$8e-3            & 82.20$\pm$11e-3            & 95.23$\pm$5e-3         \\
MULDA-m         & 92.59$\pm$10e-3           & 82.17$\pm$6e-3             & 95.12$\pm$4e-3         \\
MvMDA           & 92.65$\pm$8e-3            & 80.50$\pm$13e-3            & 93.78$\pm$9e-3         \\
OGMA            & 95.01$\pm$5e-3            & 86.00$\pm$10e-3            & 96.09$\pm$6e-3         \\
OMLDA           & 94.98$\pm$5e-3            & 86.85$\pm$10e-3            & 95.71$\pm$6e-3         \\
OMvMDA          & 94.71$\pm$7e-3            & 82.28$\pm$10e-3            & 95.99$\pm$6e-3         \\
M$^2$CPC-p      & 94.83$\pm$1.1             & 86.44$\pm$1.1              & -                  \\
MV-GWCN-1       & 95.25$\pm$1.3             & 87.83$\pm$0.8              & 96.45$\pm$1.4          \\
MV-GWCN-2       & \textbf{96.23$\pm$0.7}    & \textbf{88.46$\pm$1.1}     & \textbf{97.21$\pm$0.7} \\ \hline
\end{tabular}
\end{table}

In the last experiment, we develop new network architectures based on several conventional graph convolutional networks for applying to multi-view datasets. The architecture of these developed networks, that are named with a prefix 'M' in Table \ref{table9} to show this extension, is simple. Similar to MV-GWCN, each modality is represented through K layers using a specific graph convolutional network, and then embedded features in each modality are fused into a joint layer to have better representation based on all modalities.\par

In this way, we evaluate the abilities of various graph convolutional networks in both feature mapping in each modality as well as cross-modal feature fusion, and then compare these networks with MV-GWCN, which takes advantage of graph wavelet bases abilities. \par

The classification results of various graph convolutional networks on multi-view datasets are shown in Table \ref{table9}. The results reported in this table demonstrate the superiority of applying wavelet bases on GNNs with similar modality fusion idea. These results confirm the effectiveness of applying wavelet bases in simultaneously representing each modality and utilizing correlation among all modalities efficiently. \par

\begin{table}[!t]
\caption{Classification accuracies on multi-view datasets (Mean $\pm$ Standard Deviation)}
\label{table9}
\centering
\begin{tabular}{lccc}
\hline
\textbf{Method} & \textbf{Caltech101-7} & \textbf{Caltech101-20} & \textbf{MNIST}     \\ \hline
M-ChebyNet      & 91.49$\pm$0.3             & 85.53$\pm$0.4              & 95.50$\pm$0.8          \\
M-GAT           & 93.59$\pm$0.4             & 82.59$\pm$0.5              & 96.00$\pm$0.4          \\
M-GIN           & 92.51$\pm$0.7             & 84.27$\pm$0.3              & 95.25$\pm$0.6          \\
M-GCN           & 91.15$\pm$0.3             & 82.80$\pm$0.6              & 94.24$\pm$0.3          \\
M- GNN-ARMA     & 94.89$\pm$1.1             & 86.37$\pm$0.5              & 95.75$\pm$0.3          \\
MV-GWCN-1       & \textbf{95.25$\pm$1.3}    & \textbf{87.83$\pm$0.8}     & \textbf{96.45$\pm$1.4} \\ \hline
\end{tabular}
\end{table}

\section{Conclusions}
\label{sec6.conclusions}

Extending the convolutional neural networks to multimodal geometrically structured and/or graph-based data is an important problem that has been rarely addressed to the best of authors knowledge. This paper introduced a novel graph convolutional neural network based on wavelet bases to learn the representation of multimodal graph-based data in the spectral domain. Compared with Fourier bases used in spectral GNNs, wavelet bases more effectively represent the feature vectors in each modality utilizing scaling parameter. Besides, due to the ability of the Chebyshev polynomials used in approximating wavelet bases without requiring the costly eigen-decomposition and sparsity structure of obtained wavelet bases, computations are performed efficiently. \par

Proposed M-GWCN simultaneously provides intra-modal localization by applying multi-scaled graph wavelet convolution, and as an alignment stage estimates the cross-modal correlations between various modalities. We also introduced two additional particular versions of the proposed network for conducting unimodal and multi-view tasks.\par

The proposed network evaluated on both unimodal explicit graph-based data sets as well as multimodal implicit graph-based data. Extensive experiments demonstrated that M-GWCN outperforms state-of-the-art GNNs, including unimodal and multimodal cases, without any prior knowledge.\par

To the best of our knowledge, our proposed M-GWCN model is the first work in developing graph convolutional networks for multimodal graph-based data in the practical scenarios. \par

According to the efficiency, generality, and flexibility of the spatial methods, analyzing the multimodal graph-based data using a spatial network can be a valuable work for future. \par

Since multimodal data have a wide range of applications, developing multimodal graph-based networks for other tasks, including cross-modal retrieval, multimodal clustering, domain adaptation, etc., can also be our other future works.

\bibliographystyle{IEEEtran}

\bibliography{mybibfile}

\end{document}